\def\year{2020}\relax
\documentclass[letterpaper]{article} 
\usepackage{arXiv_V2_aaai_sty_mods}  
\usepackage{times}  
\usepackage{helvet} 
\usepackage{courier}  
\usepackage[hyphens]{url}  
\usepackage{graphicx} 
\usepackage[utf8]{inputenc} 
\usepackage[T1]{fontenc}    
\usepackage{url}            
\usepackage{booktabs}       
\usepackage{amsfonts}       
\usepackage{nicefrac}       
\usepackage{multirow}
\usepackage{latexsym}
\usepackage{svg}
\usepackage{wrapfig}
\usepackage{rotating}

\usepackage{array}

\usepackage{amsmath}
\usepackage{float}
\usepackage{lipsum}     
\usepackage{xargs}      
\usepackage{graphicx}  
\title{On Extractive and Abstractive Neural Document Summarization \\ with Transformer Language Models}
\author{Sandeep Subramanian$^{1,2,3,}$\thanks{Equal contribution, order determined by coin flip} , Raymond Li$^{1,*}$, Jonathan Pilault$^{1,2,4,*}$, Christopher Pal$^{1,2,4,5}$ \\
$^{1}$Element AI, $^{2}$Montréal Institute for Learning Algorithms, $^{3}$Université de Montréal, \\$^{4}$École Polytechnique de Montréal, $^{5}$Canada CIFAR AI Chair \\
\texttt{$^{1}$\{jonathan.pilault\}@elementai.com}} 
\begin{document}

\maketitle

\begin{abstract}
We present a method to produce abstractive summaries of long documents that exceed several thousand words via neural abstractive summarization. We perform a simple extractive step before generating a summary, which is then used to condition the transformer language model on relevant information before being tasked with generating a summary. We show that this extractive step significantly improves summarization results. We also show that this approach produces more abstractive summaries compared to prior work that employs a copy mechanism  while still achieving higher rouge scores.
\textit{Note: The abstract above was not written by the authors, it was generated by one of the models presented in this paper based on an earlier draft of this paper.}
\end{abstract}

\section{Introduction}
\label{sec:intro}
Language models (LMs) are trained to estimate the joint probability of an arbitrary sequence of words or characters using a large corpus of text. They typically factorize the joint distribution of tokens $p(x_1, x_2 \ldots x_n)$ into a product of conditional probabilities $\prod_{i}^{n} p(x_i|x_{<i})$. It is possible to use n-gram based models to estimate these conditional probabilities via counts, relying on Markovian assumptions. However, Markovian assumptions and the curse of dimensionality make it harder for n-gram LMs to model long range dependencies and learn smooth functions that can learn similarities between words in the vocabulary. This has led to a preference for recurrent or feed-forward neural language models \cite{bengio2003neural,mikolov2010recurrent} in recent years due to to their ability to learn expressive conditional probability distributions \cite{radford2019language}.

\begin{figure}[htb]
    \center{\includegraphics[scale=0.35] 
    {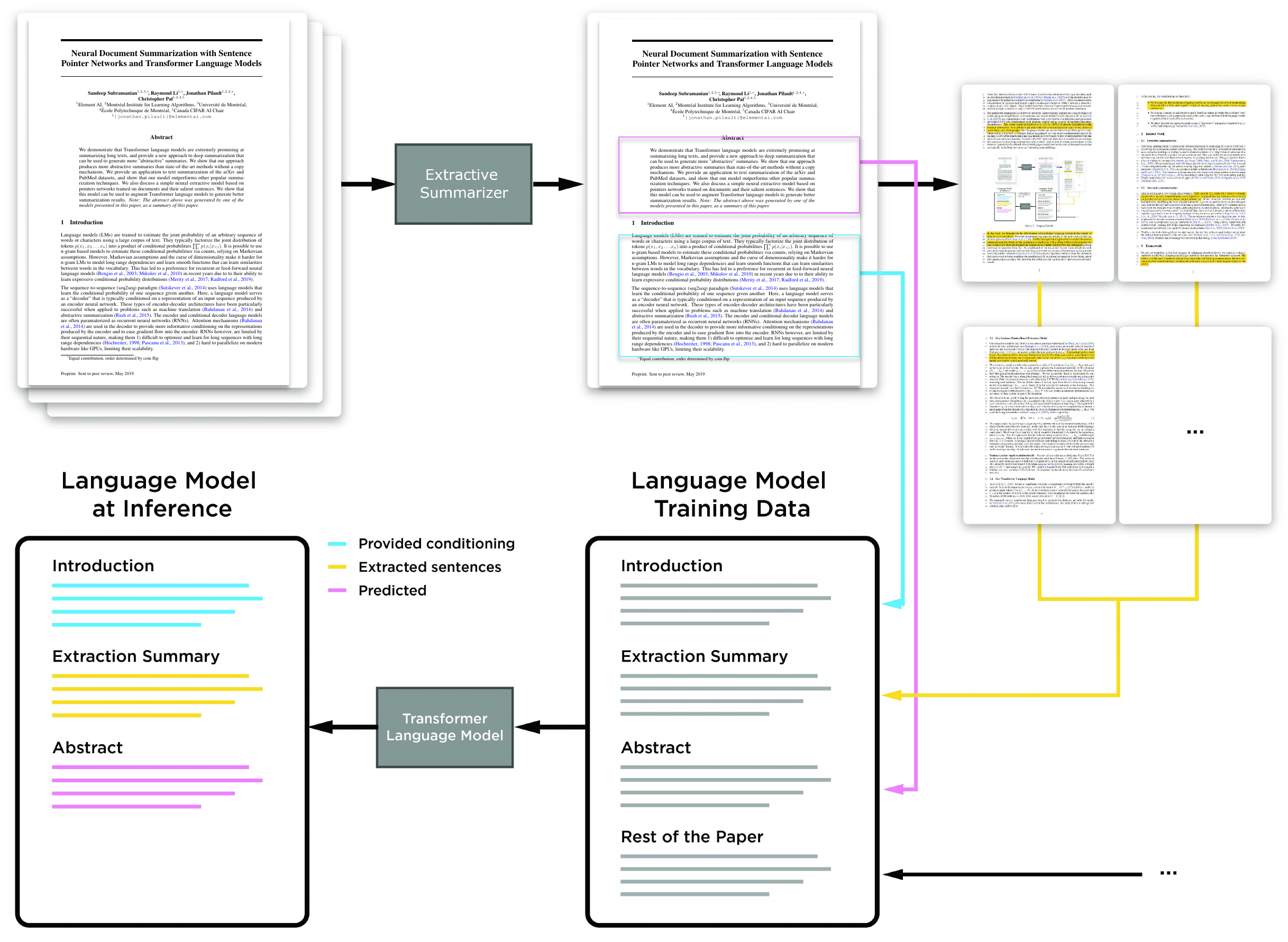}}
    \caption{\label{fig:model} \small Proposed model for abstractive summarization of a scientific article. An older version of this paper is shown as the reference document. First, a sentence pointer network extracts important sentences from the paper. Next, these sentences are provided along with the whole scientific article to be arranged in the following order: Introduction, extracted Sentences, abstract \& the rest of the paper. A transformer language model is trained on articles organized in this format. During inference, the introduction and the extracted sentences are given to the language model as context to generate a summary. In domains like news and patent documents, the introduction is replaced by the entire document. }
\end{figure}

The sequence-to-sequence (seq2seq) paradigm \cite{sutskever2014sequence} uses language models that learn the conditional probability of one sequence given another. Here, a language model serves as a ``decoder'' that is typically conditioned on a representation of an input sequence produced by an encoder neural network. These types of encoder-decoder architectures have been particularly successful when applied to problems such as machine translation \cite{bahdanau2014neural} and abstractive summarization \cite{Rush2015AbsAttn}. The encoder and conditional decoder language models are often parameterized as recurrent neural networks (RNNs). Attention mechanisms \cite{bahdanau2014neural} are used in the decoder to provide more informative conditioning on the representations produced by the encoder and to ease gradient flow into the encoder. RNNs however, are limited by their sequential nature, making them 1) difficult to optimize and learn for long sequences with long range dependencies \cite{hochreiter1998vanishing,pascanu2013difficulty}, and 2) hard to parallelize on modern hardware like GPUs, limiting their scalability.

There has therefore been a recent shift towards feedforward architectures for sequential data, such as convolutional models \cite{kalchbrenner2016neural,van2016wavenet,gehring2017convolutional} or fully attentive models popularized by architectures known as transformers \cite{DBLP:journals/corr/VaswaniSPUJGKP17}. These techniques have a logarithmic or constant path length (as opposed to linear in RNNs) between a network's output and any of its inputs, making gradient flow much easier, thereby opening up the possibility of learning very long term dependencies.

The abstractive summarization of news or scientific papers typically requires encoding and generating hundreds or thousands of words. Recent work by \cite{radford2019language} (GPT-2) has demonstrated that transformers with a large receptive field and trained on a lot of data yield language models that are capable of capturing long range dependencies.

If one is interested in creating a coherent, high-quality summary of long documents, such GPT-like architectures possess many desirable properties. Their results also show that unconditional language models can implicitly learn to perform summarization or machine translation as a consequence of the data on which it is trained. If the data is formatted sequentially into different aspects of a document (introduction, body, summary), each divided by ``tl;dr'', the model can be coaxed to generate one of these aspects. For example, the model can be made to solve a summarization task by presenting it similarly formatted data at test time; i.e. a document's introduction and body followed by ``tl;dr`` that will generate an abstract from a language model conditioned on this context.

In this work, we take this idea a step further by doing away with the sequence-to-sequence paradigm and formatting data for abstractive summarization in a manner that transformer language models can make use of all of the available data present in the documents and their summaries (akin to language model pre-training on mono-lingual data in machine translation \cite{gulcehre2015using}). Specifically, we use a single GPT-like Transformer LM (TLM) trained on documents followed by their summaries. During inference, we generate from the LM, conditioned on the document (see figure \ref{fig:model}). Unlike most previous approaches to neural abstractive summarization, we do not use a seq2seq formulation with an explicit encoder and decoder for word generation. We split the task in two: an extractive step and an abstractive step \cite{chen2018fast,gehrmann2018bottom}. To deal with extremely long documents that exceed several thousand words, we first perform sentence extraction using two different hierarchical document models - one based on pointer networks \cite{vinyals2015pointer}, similar to the variant proposed in \cite{chen2018fast} and the other based on a sentence classifier \cite{nallapati2017summarunner}. This extracts important sentences from the document (described in section \ref{sec:extractive_model}) that can be used to better condition the transformer LM on relevant information before being tasked with generating a summary. We show that this extractive step significantly improves summarization results.

The contributions of this work are two fold:

\begin{itemize}
    \item We demonstrate that transformer language models are surprisingly effective at summarizing long scientific articles and outperform typical seq2seq approaches, even without a copy mechanism.
    \item We show that our approach produces more ``abstractive'' summaries compared to prior work that employs a copy mechanism \cite{asee2017pointer} while still achieving higher ROUGE scores.
\end{itemize}

\section{Related Work}
Automatic summarization systems seek to condense the size of a piece of text while preserving most of its important information content and meaning. The earliest attempts at automatic summarization focused on extractive techniques, which find words or sentences in a document that capture its most salient content.
In the past, various similarity scores based on specific sentence features (keywords, position, length, frequency, linguistic) and metrics (structure-based, vector-based and graph-based) were employed to estimate salience \cite{steinberger2004using,erkan2004lexrank} between a sentence in a document and its reference summary. More recently, with advances in distributed representations of words, phrases and sentences, researchers have proposed to use these to compute similarity scores. Such techniques were further refined by \cite{nallapati2016classify,cheng2016neural,chen2018fast} with encoder-decoder architectures - the representations learned by the encoder are used to choose the most salient sentences.
\cite{cheng2016neural} and \cite{nallapati2016classify} trained encoder-decoder neural networks as a binary classifier to determine if each sentence in a document should belong to the extractive summary or not. \cite{nallapati2016abstractive} also present an alternative that can pick an unordered set of sentences from the source document to assemble an extractive summary. \cite{chen2018fast} use a pointer network \cite{vinyals2015pointer} to sequentially pick sentences from the document that comprise its extractive summary.

Human summarizers have four common characteristics. They are able to (1) interpret a source document, (2) prioritize the most important parts of the input text, (3) paraphrase key concepts into coherent paragraphs and (4) generate diverse output summaries. While extractive methods are arguably well suited for identifying the most relevant information, such techniques may lack the fluency and coherency of human generated summaries. Abstractive summarization has shown the most promise towards addressing points (3) and (4) above. Abstractive generation may produce sentences not seen in the original input document. Motivated by neural network success in machine translation experiments, the attention-based encoder decoder paradigm has recently been widely studied in abstractive summarization \cite{Rush2015AbsAttn,nallapati2016abstractive,chopra2016abstractive}. By dynamically accessing the relevant pieces of information based on the hidden states of the decoder during generation of the output sequence, the model revisits the input and attends to important information. The advantages of extractive, abstractive and attention-based models were first combined in \cite{gu2016copy} with a copy mechanism for out-of-vocabulary words present in the source document. Similarly, \cite{asee2017pointer} used the attention scores to calculate the probability of generating vs copying a word. A coverage mechanism was also added to penalize the attention score of previously attended words, diminishing the model's tendency to repeat itself.

\section{Framework}
Our model comprises two distinct and independently trainable components 1) a hierarchical document representation model that either points to or classifies sentences in a document to build an extractive summary 2) a transformer language model that conditions on the extracted sentences as well as a part of or the entire document.

\subsection{Extractive Models}
\label{sec:extractive_model}
We describe the two neural extractive models used in this work in this section. \\ \\
\textbf{Hierarchical Seq2seq Sentence Pointer}
Our extractive model is similar to the sentence pointer architecture developed by \cite{chen2018fast} with the main difference being the choice of encoder. We use a hierarchical bidirectional LSTM encoder with word and sentence level LSTMs while \cite{chen2018fast} use a convolutional word level encoder for faster training and inference. The decoder is in both cases is an LSTM.

The extractive model considers the document as a list of $N$ sentences $D = \left( S_1, \ldots, S_N \right)$, and each sentence as a list of tokens. We are given a ground-truth extracted summary of $M$ sentences $\left(S_{i_1}, \ldots, S_{i_M}\right)$, where the $i_1 < \ldots < i_M$ are the indices of the extracted sentences.
The procedure to determine ground-truth extraction targets are identical to previous work - finding two sentences in the document that have the highest ROUGE score with each sentence in the summary.

We use an encoder-decoder architecture for this extractor.
The encoder has a hierarchical structure that combines a token and sentence-level RNN. First, the ``sentence-encoder'' or token-level RNN is a bi-directional LSTM \cite{Hochreiter:1997:LSM:1246443.1246450} encoding each sentence. The last hidden state of the last layer from the two directions produces sentence embeddings: $(\mathbf{s}_1, \ldots, \mathbf{s}_N)$, where $N$ is the number of sentences in the document. The sentence-level LSTM or the ``document encoder'', another bi-directional LSTM, encodes this sequence of sentence embeddings to produce document representations: $(\mathbf{d}_1, \ldots, \mathbf{d}_N)$.

The decoder is an autoregressive LSTM taking the sentence-level LSTM hidden state of the previously extracted sentence as input and predicting the next extracted sentence.
Let $i_t$ the index of the previous extracted sentence at time step $t$. The input to the decoder is $\mathbf{s}_{i_t}$, or a zero vector at time-step $t=0$.
The decoder's output is computed by an attention mechanism from the decoder's hidden state $\mathbf{h}_t$ over the document representations $(\mathbf{d}_1, \ldots, \mathbf{d}_N)$.
We used the dot product attention method from \cite{luong2015effective}.
The attention weights $\mathbf{a}_t$ produce a context vector $\mathbf{c}_t$, which is then used to compute an attention aware hidden state $\mathbf{\Tilde{h}}_t$.
Following the input-feeding approach from \cite{luong2015effective}, the attention aware hidden state $\mathbf{\Tilde{h}}_t$ is concatenated to the input in the next time step, giving the following recurrence
$\mathbf{h}_t = \text{LSTM}\left([
           \mathbf{s}^T_{i_t} \;
           \mathbf{\Tilde{h}}^T_{t-1}
         ]^T, \mathbf{h}_{t-1}\right)$, with
\begin{flalign}
    \mathbf{\Tilde{h}}_t = \mathbf{W}_{\Tilde{h}} \begin{bmatrix}
       \mathbf{c}_{t} \\
       \mathbf{h}_t
    \end{bmatrix},
     \; \;
     \mathbf{c}_t = \sum_{i=1}^N a_t(i)\mathbf{d}_i, 
     \; \;
    \alpha_{t}(i) = \mathbf{d}_i^T \mathbf{h}_t, \; \;  
    \\
    a_t(i) = \frac{\exp\left(\alpha_{t}(i)\right)}{\sum_{i'} \exp\left(\alpha_{t}(i')\right)}, \; \; 
    \text{for} \; \; i=1..N .
\end{flalign}

The attention weights $\mathbf{a}_t$ are used as output probability distribution over the document sentences, of the choice for the next extracted sentence.
We choose the convention to signal the end of the extraction by putting the same index twice in a row.
Thus, the input to the decoder is the following sequence:
    $\mathbf{0}, \mathbf{s}_{i_1}, ..., \mathbf{s}_{i_M}$
, and the target:
    $i_1, ..., i_M, i_M$
, where $M$ is the length of the ground-truth extracted summary and both sequences have $M+1$ elements.
The model is trained to minimize the cross-entropy of picking the correct sentence at each decoder time step. At inference, we use beam-search to generate the extracted summary. 

\paragraph{Sentence Classifier}
As with the pointer network, we use a hierarchical LSTM to encode the document and produce a sequence of sentence representations $\mathbf{d}_1, ..., \mathbf{d}_N$ where $N$ is the number of sentences in the document.
We compute a final document representation as follows:
\begin{equation}
    \mathbf{d} = \tanh\left( \mathbf{b}_d + \mathbf{W}_d \frac{1}{N} \sum _{i=1}^N \mathbf{d}_i \right)
\end{equation}
where $\mathbf{b}_d$ and $\mathbf{W}_d$ are learnable parameters.
Finally, the probability of each sentence belonging to the extractive summary is given by:
\begin{align}
    o_i = & \sigma \left( \mathbf{W}_o \begin{bmatrix}
           \mathbf{d}_{i} \\
           \mathbf{d}
         \end{bmatrix}  + \mathbf{b}_o \right)
\end{align}
where $\sigma$ is the sigmoid activation function. The model is trained to minimize the binary cross-entropy loss with respect to the sentences in the gold-extracted summary.

\paragraph{Model Details} The model uses word embeddings of size $300$. The token-level LSTM (sentence encoder), sentence-level LSTM (document encoder) and decoder each have $2$ layers of $512$ units and a dropout of $0.5$ is applied at the output of each intermediate layer. We trained it with Adam, a learning rate $0.001$, a weight decay of $10^{-5}$, and using batch sizes of $32$. We evaluate the model every $200$ updates, using a patience of $50$. At inference, we decode using beam search with a beam size of $4$ for the pointer model and pick the $k$ most likely sentences from the sentence classifier, where $k$ is the average number of sentences in the summary across the training dataset.

\subsection{Transformer Language Models (TLM)}
\label{sec:gpt}
Instead of formulating abstractive summarization as a seq2seq problem using an encoder-decoder architecture, we only use a single transformer language model that is trained \textit{from scratch}, with appropriately ``formatted'' data (see figure \ref{fig:model}, we also describe the formatting later in this section).

We use a transformer \cite{DBLP:journals/corr/VaswaniSPUJGKP17} language model (TLM) architecture identical to \cite{radford2019language}. Our model has 220M parameters with 20 layers, 768 dimensional embeddings, 3072 dimensional position-wise MLPs and 12 attention heads. The only difference in our architectures (to our knowledge) is that we do not scale weights at initialization. We trained the language model for 5 days on 16 V100 GPUs on a single Nvidia DGX-2 box. We used a linear ramp-up learning rate schedule for the first $40,000$ updates, to maximum learning rate of $2.5 \times e^{-4}$ followed by a cosine annealing schedule to 0 over the next $200,000$ steps with the Adam optimizer. We used mixed-precision training \cite{micikevicius2017mixed} with a batch size of 256 sequences of 1024 tokens each.

In order to get an unconditional language model to do abstractive summarization, we can use the fact that LMs are trained by factorizing the joint distribution over words autoregressively. We organized the training data for the LM such that the ground-truth summary \textit{follows} the information used by the model to generate a system summary. This way, we model the joint distribution of document and summary during training, and sample from the conditional distribution of summary given document at inference. 

When dealing with extremely long documents that may not fit into a single window of tokens seen by a transformer language model, such as an entire scientific article, we use its introduction as a proxy for having enough information to generate an abstract (summary) and use the remainder of the paper as in domain language model training data (Fig \ref{fig:model}). In such cases, we organize the arXiv and PubMed datasets as follows: 1) paper introduction 2) extracted sentences from the sentence pointer model 3) abstract 4) rest of the paper. On other datasets, the paper introduction would be the entire document and there would no rest of the paper. This ensures that at inference, we can provide the language model the paper introduction and the extracted sentences as conditioning to generate its abstract. We found that using the ground truth extracted sentences during training and the model extracted sentences at inference performed better than using the model extracted sentences everywhere.

We use a special token to indicate the start of the summary and use it at test time to signal to the model to start generating the summary. The rest of the article is provided as additional in-domain training data for the LM.
The entire dataset is segmented into non-overlapping examples of $1,024$ tokens each. We use ``topk'' sampling at inference \cite{fan2018hierarchical,radford2019language}, with $k=30$ and a softmax temperature of 0.7 to generate summaries.
\begin{figure}[htb]
    \center{\includegraphics[scale=0.43]
    {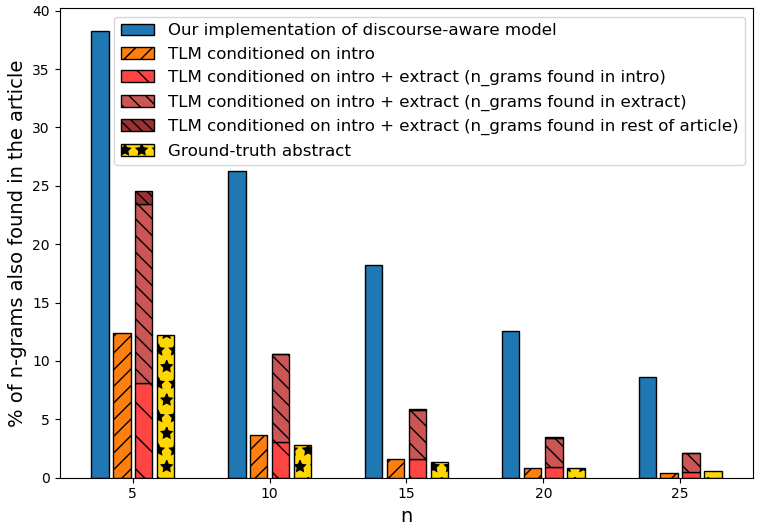}}
    \caption{\label{fig:abstractiveness} $n$-gram overlaps between the abstracts generated by different models and the input article on the arXiv dataset. We show in detail which part of the input was copied for our TLM conditioned on intro + extract.}
\end{figure}

\section{Results and Analysis}
\subsection{Experimental setup}
\label{sec:experimental_setup}
\paragraph{Datasets}
We experiment with four different large-scale and long document summarization datasets - arXiv, PubMed \cite{discourse/corr/abs-1804-05685}, bigPatent \cite{sharma2019bigpatent} and Newsroom \cite{grusky2018newsroom}. Statistics are reported in Table \ref{tab:dataset}.

\begin{table}[ht]
\centering
\caption{Statistics from \cite{sharma2019bigpatent} for the datasets used in this work - The number of document/summary pairs, the ratio of the number of words in the document to the abstract and the number of words in the summary and document.}
\label{tab:dataset}
\small
\begin{tabular}{|l|c|c|c|c|}
	\hline 
		Dataset & \#Documents & \shortstack{Comp \\ Ratio} & \shortstack{Sum \\ Len} & \shortstack{Doc \\ Len}  \\
		\hline
        arXiv & 215,913 & 39.8 & 292.8 & 6,913.8 \\ 
        PubMed & 133,215 & 16.2 & 214.4 & 3,224.4 \\
        Newsroom & 1,212,726 & 43.0 & 30.4 & 750.9 \\
        BigPatent & 1,341,362 & 36.4 & 116.5 & 3,572.8 \\
    \hline
\end{tabular}
\end{table}

\paragraph{Data preprocessing} Both our extractive and abstractive models use sub-word units computed using \emph{byte pair encoding} \cite{sennrich2015neural} with $40,000$ replacements. To address memory issues in the sentence pointer network, we only keep $300$ sentences per article, and $35$ tokens per sentence.

\paragraph{Evaluation} We evaluate our method using full-length F-1 ROUGE scores \cite{rouge} and re-used the code from \cite{discourse/corr/abs-1804-05685} for this purpose. All ROUGE numbers reported in this work have a 95\% confidence interval of at most 0.24.

\paragraph{Comparison} We compare our results to several previously proposed extractive and abstractive models. All prior results reported on the arXiv and Pubmed benchmark are obtained from \cite{discourse/corr/abs-1804-05685}. Similarly, prior results for the BigPatent dataset are obtained from \cite{sharma2019bigpatent} and Newsroom from \cite{grusky2018newsroom} and \cite{mendes2019jointly}. These methods include \emph{LexRank} \cite{erkan2004lexrank}, \emph{SumBasic} \cite{vanderwende2007beyond}, \emph{LSA} \cite{steinberger2004using}, \emph{Attention-Seq2Seq} \cite{nallapati2016abstractive,chopra2016abstractive}, \emph{Pointer-Generator Seq2Seq} \cite{asee2017pointer},
\emph{Discourse aware}, which is a hierarchical extension to the pointer generator model, \cite{discourse/corr/abs-1804-05685}, \emph{Sent-rewriting} \cite{chen2018fast}, \emph{RNN-Ext} \cite{chen2018fast}, \emph{Exconsumm} \cite{mendes2019jointly}.

\subsection{Discussion}
\begin{figure}[htb]
    \center{\includegraphics[scale=0.43]
    {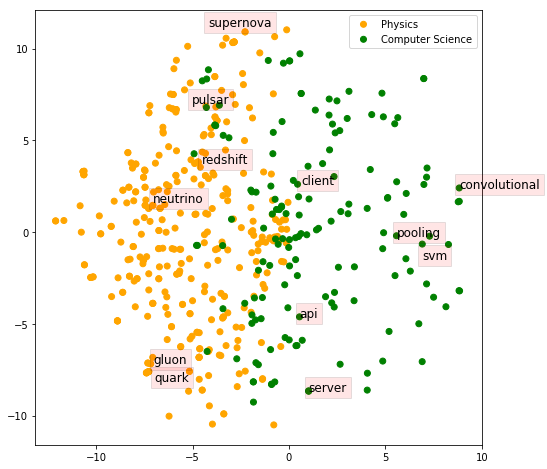}}
    \caption{\label{fig:t-sne} t-sne visualization of the TLM-learned word embeddings. The model appears to partition the space based on the broad paper categoty in which it frequently occurs.}
\end{figure}

We present our main results on summarizing arXiv and PubMed papers in tables \ref{tab:arxiv_abstract}, \ref{tab:pubmed_abstract}. Our extractive models are able to outperform previous extractive baselines on both the arXiv and Pubmed datasets.
Our TLM conditioned on the extractive summary produced by our best extractive model (TLM-I+E (G,M)) outperforms prior abstractive/mixed results on the arXiv, Pubmed and bigPatent datasets, except on ROUGE-L. 
On Newsroom, we do better than the only other abstractive model (Seq2Seq with attention) by a massive margin and achieve better performance than the pointer generator even on the abstractive and mixed which their model should be better suited to since it has a copy mechanism. The Exconsumm model \cite{mendes2019jointly}  however, which is primarily an extractive model does better on this dataset. We suspect the poor ROUGE-L result is due to the absence of a copy mechanism that makes it hard to get \textit{exact} large n-gram matches. Figure \ref{fig:abstractiveness} further supports this hypothesis, it is evident that a model with a copy mechanism is often able to copy even upto 25-grams from the article. Further, \cite{graham2015re} finds that ROUGE-L is poorly correlated with human judgements when compared to ROUGE-1,2,3. In table \ref{tab:select_papers} and Table \ref{tab:arxiv_random_set}, we present qualitative results of abstracts of notable papers in our field and of our TLM conditioned on the introductions and extracted summaries of a random example from the arXiv test set. Table \ref{tab:qualitative_intro_extract} shows similar qualitative examples on the Newsroom dataset. Tables \ref{tab:arxiv_abstract}, \ref{tab:pubmed_abstract} and \ref{tab:patent_abstract} also provide different train / test settings for our TLM conditioned on extracted sentences. We show a performance upper bound conditioning the Transformer LM on oracle / ground-truth extracted sentences at both train and test time (TLM-I+E (G,G)). We also experiment with using either the ground-truth extracted sentences (TLM-I+E (G,M)) or the model extracted sentences (TLM-I+E (M,M)) during training and find that latter slightly impairs performance. Finally, figure \ref{fig:t-sne}, presents a visualization of the word embeddings learned by our TLM.

\begin{table}[ht]
\centering
\caption{Summarization results on the arXiv dataset. Previous work results from \cite{discourse/corr/abs-1804-05685}. The following lines are a simple baseline Lead-10 extractor and the pointer and classifier models. Our transformer LMs (TLM) are conditioned either on the Introduction (I) or along with extracted sentences (E) either from ground-truth (G) or model (M) extracts.}
\label{tab:arxiv_abstract}
\small
\begin{tabular}{|l|c|cccc|}
	\hline 
		\multirow{2}*{Model} & \multirow{2}*{Type} 
		&  \multicolumn{4}{c|}{ROUGE}  \\
        & & 1 & 2 & 3 & L  \\ \hline
        \multicolumn{6}{|c|}{\textbf{Previous Work}} \\
        \hline
        SumBasic  & Ext & 29.47 & 6.95 & 2.36 & 26.3  \\
        LexRank  & Ext & 33.85 & 10.73 & 4.54 & 28.99  \\
        LSA  & Ext & 29.91 & 7.42 & 3.12 & 25.67  \\
        \hline
        Seq2Seq & Abs & 29.3 & 6.00 & 1.77 & 25.56  \\
        Pointer-gen & Mix & 32.06 & 9.04 & 2.15 & 25.16 \\
        Discourse & Mix & 35.80 & 11.05 & 3.62 & 31.80 \\
        \hline
        \multicolumn{6}{|c|}{\textbf{Our Models}} \\
        \hline
        Lead-10 & Ext & 35.52 & 10.33 & 3.74 & 31.44 \\
        Sent-CLF & Ext & 34.01 & 8.71 & 2.99 & 30.41 \\ 
        Sent-PTR & Ext & \underline{42.32} & \underline{15.63} & \underline{7.49} & \underline{38.06} \\
        \hline
        TLM-I & Abs & 39.65 & 12.15 & 4.40 & 35.76 \\ 
        TLM-I+E (M,M) & Mix & 41.15 & 13.98 & 5.63 & 37.40 \\ 
        TLM-I+E (G,M) & Mix & \textbf{41.62} & \textbf{14.69} & \textbf{6.16} & \textbf{38.03}  \\ 
        \hline
        \multicolumn{6}{|c|}{\textbf{Oracle}} \\
        \hline
        Gold Ext & Oracle & 44.25 & 18.17 & 9.14 & 35.33  \\
        TLM-I+E (G,G) & Oracle & 46.40 & 18.15 & 8.71 & 42.27 \\ 
    \hline
\end{tabular}
\end{table}

\begin{table}[h!]
\caption{\small Qualitative Results - News articles and our model generated summaries on the NewsRoom dataset}
\begin{center}
\scriptsize
\begin{tabular}{|m{7.9cm}|}
\hline
\textbf{Document} --- A new plan from the government of the Philippines would offer free wireless internet to people across the country while also likely eating into the annual revenue of the nations telecoms. Bloomberg reports that the Philippines government plans to roll-out its free Wi-Fi services to roughly half of the countrys municipalities over the next few months and the country has its sights set on nationwide coverage by the end of 2016. The free wireless internet service will be made available in public areas such as schools, hospitals, airports and parks, and is expected to cost the government roughly \$32 million per year. [...] \\
\hline
\textbf{Abstractive} --- : The government is reportedly considering a nationwide service plan to give free Wi-Fi access to rural areas. \\
\hline
\textbf{Mixed} --- The government of the Philippines is considering a new plan to provide free wireless internet to the nation's largest cities and towns.  \\
\hline
\textbf{Extractive} --- The new plan will include free wireless internet to residents across the country while also probably eating into the annual revenue of the country's telecoms. \\
\hline
\textbf{Document} --- (CBS) - Controversy over a new Microsoft patent has people questioning whether or not the intention has racist undertones. CNET reported that Microsoft has been granted a U.S. patent that will steer pedestrians away from areas that are high in crime. [...] \\
\hline
\textbf{Absractive Summary} --- The new Microsoft patent claims a device could provide pedestrian navigation directions from a smartphone. \\
\hline
\textbf{Mixed Summary} Microsoft won a U.S. patent for a new way to steer pedestrians out of areas that are high in crime \\
\hline
\end{tabular}
\end{center}
\label{tab:qualitative_intro_extract}
\end{table}

\begin{table}[ht]
\begin{center}
\caption{Summarization results on the PubMed dataset. Previous work results from \cite{discourse/corr/abs-1804-05685}. The following lines are a simple baseline Lead-10 extractor and the pointer and classifier models. Our transformer LMs (TLM) are conditioned either on the Introduction (I) or along with extracted sentences (E) either from ground-truth (G) or model (M) extracts.}
\label{tab:pubmed_abstract}
\small
\begin{tabular}{|l|c|cccc|}
	\hline 
		\multirow{2}*{Model} & \multirow{2}*{Type}
		&  \multicolumn{4}{c|}{ROUGE}  \\
        & & 1 & 2 & 3 & L  \\ \hline
        \multicolumn{6}{|c|}{\textbf{Previous Work}} \\
        \hline
        SumBasic  & Ext & 37.15 & 11.36 & 5.42 & 33.43  \\
        LexRank  & Ext & 39.19 &  13.89 &  7.27 &  34.59  \\ 
        LSA  & Ext & 33.89 & 9.93 & 5.04 & 29.70 \\ 
        \hline
        Seq2seq & Abs & 31.55 & 8.52 & 7.05 & 27.38  \\
        Pointer-gen & Mix & 35.86 & 10.22 & 7.60 & 29.69 \\
        Discourse & Mix & 38.93 & 15.37 & \textbf{9.97} & 35.21 \\
        \hline
        \multicolumn{6}{|c|}{\textbf{Our Models}} \\
        \hline
        Lead-10 & Ext & 37.45 & 14.19 & 8.26 & 34.07 \\
        Sent-CLF & Ext & \underline{45.01} & \underline{19.91} & \underline{12.13} & \underline{41.16} \\ 
        Sent-PTR & Ext & 43.30 & 17.92 & 10.67 & 39.47 \\
        \hline
        TLM-I & Abs & 37.06 & 11.69 & 5.31 & 34.27 \\ 
        TLM-I+E (G,M)& Mix & \textbf{42.13} & \textbf{16.27} & 8.82 & \textbf{39.21} \\ 
        \hline
        \multicolumn{6}{|c|}{\textbf{Oracle}} \\
        \hline
        Gold Ext & Oracle & 47.76 & 20.36 & 11.52 & 39.19  \\
        TLM-I+E (G,G) & Oracle & 46.32 & 20.15 & 11.75 & 43.23 \\ 
    \hline
\end{tabular}
\end{center}
\end{table}

\begin{table}[ht]
\begin{center}
\caption{Summarization results on the bigPatent dataset. Previous work results from \cite{sharma2019bigpatent}. Our transformer LMs (TLM) are conditioned on the whole document or additionally with extracted sentences (E) either from ground-truth (G) or model (M) extracts.}
\label{tab:patent_abstract}
\small
\begin{tabular}{|l|c|ccc|}
	\hline 
		\multirow{2}*{Model} & \multirow{2}*{Type} 
		&  \multicolumn{3}{c|}{ROUGE}  \\
        & & 1 & 2 & L  \\ \hline
        \multicolumn{5}{|c|}{\textbf{Previous Work}} \\
        \hline
        Lead-3     & Ext & 31.27 & 8.75  & 26.18 \\
        TextRank   & Ext & 35.99 & \underline{11.14} & 29.60 \\
        SumBasic   & Ext & 27.44 & 7.08  & 23.66 \\
        LexRank    & Ext & 35.57 & 10.47 & 29.03 \\
        RNN-Ext    & Ext & 34.63 & 10.62 & 29.43 \\
        \hline
        Seq2Seq & Abs & 28.74 & 7.87 & 24.66  \\
        Pointer-gen & Mix & 30.59 & 10.01 & 25.65 \\
        Pointer-gen (Cov) & Mix & 33.14 & 11.63 & 28.55 \\
        Sent-rewriting & Mix & 37.12 & 11.87 & 32.45 \\
        \hline
        \multicolumn{5}{|c|}{\textbf{Oracle}} \\
        \hline
        Gold Ext & Oracle & 43.56 & 16.91 & 36.52  \\
        OracleFrag & Oracle & 91.85 & 78.66 & 91.85 \\
        \hline
        \multicolumn{5}{|c|}{\textbf{Our Models}} \\
        \hline
        Sent-CLF & Ext & \underline{36.20} & \underline{10.99} & \underline{31.83} \\ 
        Sent-PTR & Ext & 34.21 & 10.78 & 30.07 \\
        \hline
        TLM & Abs & 36.41 & 11.38 & 30.88  \\ 
        TLM+E (G,M) & Mix & \textbf{38.65} & \textbf{12.31} & \textbf{34.09} \\
        \hline
        TLM+E (G,G) & Oracle & 39.99 & 13.79 & 35.33 \\
    \hline
\end{tabular}
\end{center}
\end{table}

\begin{table}[h!]
\begin{center}
\caption{Summarization results on the Newsroom dataset. Previous work results from \cite{grusky2018newsroom} and \cite{mendes2019jointly}.}
\label{tab:newsroom_abstract}
\scriptsize
\begin{tabular}{|p{1.5cm}|p{0.6cm}|p{0.1cm}p{0.2cm}p{0.3cm}|p{0.1cm}p{0.2cm}p{0.3cm}|p{0.1cm}p{0.1cm}p{0.3cm}|}
	\hline 
		Model & Type & \multicolumn{3}{c|}{Extractive} & \multicolumn{3}{c|}{Mixed} & \multicolumn{3}{c|}{Abstractive} \\
		\hline
		& & \multicolumn{9}{c|}{ROUGE} \\
		\hline
		& & 1 & 2 & L & 1 & 2 & L & 1 & 2 & L  \\
		\hline
		\multicolumn{11}{|c|}{\textbf{Previous Work}} \\
		\hline
		Seq2Seq & Abs & 6.1  & 0.2  & 5.4  & 5.7  & 0.2  & 5.1  & 6.2  & 1.1  & 5.7 \\
		TextRank & Ext & 32.4 & 19.7 & 28.7 & 22.3 & 7.9 & 17.7 & 13.5 & 1.9 & 10.5 \\
		Pointer-gen & Mix & 39.1  & 27.9  & 36.2  & 25.5  & 11.0  & 21.1  & 14.7  & 2.3  & 11.4 \\
	    Lead-3  & Ext & 53.0  & 49.0  & 52.4  & 25.1  & 12.9  & 22.1  & 13.7  & 2.4 & 11.2 \\
	    Exconsumm & Mix & \textbf{68.4} & \textbf{62.9}  & \textbf{67.3}  & 31.7 & 16.1  & 27.0  & 17.1  & 3.1  & 14.1 \\
	    \hline
		\multicolumn{11}{|c|}{\textbf{Our Models}} \\
	    \hline
	    Sent-CLF & Ext & 53.0 & 47.0 & 52.1 & 26.8 & 12.6 & 23.6 & 15.4 & 2.7 & 12.8 \\
	    Sent-PTR & Ext & 60.7 & 55.2 & 59.7 & 28.9 & 14.1 & 25.1 & 15.9 & 2.8 & 13.0 \\
	    TLM & Abs & 49.8 & 39.7 & 47.4 & 27.1 & 11.6 & 22.8 & \textbf{20.4} & \textbf{6.9} & \textbf{17.1} \\ 
	    TLM+E (G,M) & Mix & 63.3 & 57.3 & 61.8 & \textbf{31.9} & \textbf{16.6} & \textbf{27.4} & 20.1 & 6.5 & 16.6 \\  
	    \hline
		\multicolumn{11}{|c|}{\textbf{Oracle}} \\
	    \hline
	    Gold Ext & Oracle & 68.1 & 64.5 & 67.3 & 40.8 & 24.6 & 34.2 & 21.9 & 5.2 & 16.3 \\
	    TLM+E (G,G) & Oracle & 78.8 & 74.0 & 77.8 & 38.6 & 22.0 & 33.6 & 24.5 & 9.6 & 20.8 \\  
    \hline
\end{tabular}
\end{center}
\end{table}

\begin{table*}[h!]
\caption{Qualitative Results --- Generated abstracts of select papers using our Intro Only TLM.}
\begin{center}
\scriptsize
\begin{tabular}{|m{7cm}|m{9cm}|}
\hline
\textbf{Generated abstract for ``the attention model paper''} \cite{bahdanau2014neural} - In this paper, we propose a neural machine translation model, which learns to align and translate a source sentence without encoding it into a fixed length vector. The proposed model allows the neural network to align and translate jointly, and in addition, predicts a target word based on the context vectors associated with these source positions. We show that the proposed model achieves significantly improved translation performance than the conventional encoder decoder neural network approach, when the sentences in the training corpus are long . Quantitative analysis reveals that the proposed model finds a linguistically plausible soft alignment between a source sentence and the target sentence. Finally, we compare the performance of the proposed model with the state-of-the-art standard neural network translation models and show that the proposed approach outperforms the conventional approach in terms of both translation performance and the quality of the translations. & \textbf{Original abstract} - Neural machine translation is a recently proposed approach to machine translation. Unlike the traditional statistical machine translation, the neural machine translation aims at building a single neural network that can be jointly tuned to maximize the translation performance. The models proposed recently for neural machine translation often belong to a family of encoder-decoders and consists of an encoder that encodes a source sentence into a fixed-length vector from which a decoder generates a translation. In this paper, we conjecture that the use of a fixed-length vector is a bottleneck in improving the performance of this basic encoder-decoder architecture, and propose to extend this by allowing a model to automatically (soft-)search for parts of a source sentence that are relevant to predicting a target word, without having to form these parts as a hard segment explicitly. With this new approach, we achieve a translation performance comparable to the existing state-of-the-art phrase-based system on the task of English-to-French translation. Furthermore, qualitative analysis reveals that the (soft-)alignments found by the model agree well with our intuition. \\
\hline
\textbf{Generated abstract for ``the GPT-2 paper''} \cite{radford201language} - Language models have been successful in creating rich representations of complex systems, including tasks such as sentiment analysis. In this paper we demonstrate their ability to generalize well across domains and tasks. We demonstrate the utility of language models by demonstrating that a variety of tasks can be modeled using language models. We demonstrate the potential of language models in downstream tasks by highlighting the ability of these models to perform a wide range of tasks in a zero shot setting.
& \textbf{Original abstract} - Natural language processing tasks, such as question answering, machine translation, reading comprehension, and summarization, are typically
approached with supervised learning on task-specific datasets. We demonstrate that language
models begin to learn these tasks without any explicit supervision when trained on a new dataset
of millions of webpages called WebText. When conditioned on a document plus questions, the answers
generated by the language model reach 55 F1 on the CoQA dataset - matching or exceeding
the performance of 3 out of 4 baseline systems without using the 127,000+ training examples.
The capacity of the language model is essential to the success of zero-shot task transfer and increasing it improves performance in a log-linear fashion across tasks. Our largest model, GPT-2,
is a 1.5B parameter Transformer that achieves state of the art results on 7 out of 8 tested language
modeling datasets in a zero-shot setting but still underfits WebText. Samples from the
model reflect these improvements and contain coherent paragraphs of text. These findings suggest
a promising path towards building language processing systems which learn to perform tasks from
their naturally occurring demonstrations.\\
\hline
\textbf{Generated abstract for ``the BERT paper''} \cite{devlin2018bert} - We develop an architecture for deep bidirectional language models that allows the representation to fuse the left and the right context of a given token. Our framework uses bidirectional transformer encoder representations that enable the model to fuse the context representations of a token. We show that our architecture achieves state-of-the-art performance on a large suite of tasks, outperforming many systems with task-specific architectures. Our results are based on a set of eleven task specific tasks that are benchmarks for a large suite of nlp tasks.
& \textbf{Original abstract} - We introduce a new language representation model called BERT, which stands for Bidirectional Encoder Representations from Transformers. Unlike recent language representation models, BERT is designed to pre-train deep bidirectional representations from unlabeled text by jointly conditioning on both left and right context in all layers. As a result, the pre-trained BERT model can be fine-tuned with just one additional output layer to create state-of-the-art models for a wide range of tasks, such as question answering and language inference, without substantial task-specific architecture modifications. 
BERT is conceptually simple and empirically powerful. It obtains new state-of-the-art results on eleven natural language processing tasks, including pushing the GLUE score to 80.5\% (7.7\% point absolute improvement), MultiNLI accuracy to 86.7\% (4.6\% absolute improvement), SQuAD v1.1 question answering Test F1 to 93.2 (1.5 point absolute improvement) and SQuAD v2.0 Test F1 to 83.1 (5.1 point absolute improvement). 
\\
\hline
\end{tabular}
\end{center}
\label{tab:select_papers}
\end{table*}

\subsection{Abstractiveness of generated abstracts}
\cite{weber2018controlling} argued that state-of-the-art abstractive summarization systems that use a copy mechanism effectively generate the summary by copying over large chunks from the article, essentially doing ``extractive'' summarization.  Following this work, we measure how much a model copies from the article by counting the proportion of $n$-grams from the generated abstract that are also found in the article.
These statistics measured on the arXiv dataset are presented in figure \ref{fig:abstractiveness}.
First, the original abstract and our TLM conditioned on the intro have small and very similar overlap fractions with the original article. A model using a pointing mechanism (we used our own implementation of the model developed by \cite{discourse/corr/abs-1804-05685})\footnote{This model achieved the following ROUGE-1, 2, 3 and L on the arXiv dataset: $41.33, 14.73, 6.80, 36.34$} copies more than our transformer model, especially for higher $n$-grams. In particular, more than $10$\% of the $20$-grams from the abstracts generated by the pointing model are also found in the article, showing that it tends to copy long sequences of words. On the other hand, our proposed model produces more ``abstractive'' summaries, demonstrating its ability to paraphrase.
Our model tends to copy longer sequences when conditioned on the introduction and the sentences from the extractor. We hypothesize that providing extracted sentences from the article that already contain a lot of words present in the reference abstract, makes the transformer's task easier, by allowing it to copy words and phrases from the extracted sentences. We find empirical evidence of this in figure \ref{fig:abstractiveness}, showing that the majority of $n$-gram copies come from the extracted sentences. For $5$-grams, close to 2/3rd of the words copied are from the extracted sentences. As the number of grams increases to $25$-grams, 4/5th of the words copied are from the extracted sentences.

\section{Conclusion}
We have demonstrated that Transformer language models can generate high-quality summaries of long sequences of text via an extractive step followed by an abstractive step. We quantitatively measure the positive impact of the extractive step, by comparing it to a abstractive model variant that only sees the input text itself. Our approach outperforms previous extractive and abstractive summarization methods on the arXiv, PubMed and bigPatent datasets and is less prone to copying entire phrases or sentences from the input text. The fluency and coherency of the sample summaries suggests that these models are ready for comprehensive human evaluation studies. As with other problem domains, we have observed that abstractive summaries generated by transformers can generate imaginary content.  We advise that such evaluations should probe multiple aspects of the summarization results including both factual correctness and coherency. We also note that for evaluating the correctness of the summaries of scientific articles and patents one must have highly trained evaluators who are willing to invest significant amounts of time to read the underlying papers and patents. Such studies could therefore require significant investments of resources.  We have also presented an upper bound on extractive + abstractive models, by conditioning the abstractive step on gold-extracted sentences. In future work, we are also interested in exploring the possibility of training the extractive and abstractive steps in an end-to-end manner. While we believe that this work is a step forward towards generating more abstractive summaries, it remains an open challenge to develop models that respect the underlying facts of the content being summarized while matching the creative ability of humans to coherently and concisely synthesize summaries. 

\bibliography{aaai}
\bibliographystyle{aaai}

\clearpage

\section{Appendix}
\subsection{Samples from the arXiv test set}
Samples from the TLM-I and TLM-I+E (G,M) models on a random example from the arXiv test set.
\begin{table}[h!]
\caption{\small Qualitative Results - Generated abstracts from our models on a random example from the test set of \cite{discourse/corr/abs-1804-05685}}
\begin{center}
\small
\begin{tabular}{|m{7.9cm}|}
\hline
\textbf{TLM conditioned on intro} ---
deep convolutional neural networks ( cnns ) have created new benchmarks in the object recognition challenge . most of these benchmark datasets contain thousands of images of interest . in addition to these datasets , several standard data - sets , such as pascal voc 2012 and the ms coco , have been used in conjunction with the same network in order to improve the object detection task . in this paper , we present a method for object localization using the deep convolutional neural networks trained for the simpler task of image - level classification . we propose localization candidates ( or bounding boxes ) which maximize classification scores for a class of interest . we explore the beam search strategy where the number of candidate bounding boxes are restricted as we traverse from the root node towards the leaf nodes . we then gradually explore smaller and smaller regions of interest till a point is reached where the classifier is not able to discriminate among the classes of interest . the detection candidates are organized in a search tree , the root node being represented by the entire test image . we then gradually explore smaller and smaller regions of interest till a point is reached where the classifier is not able to distinguish among the classes of interest . \\
\hline
\textbf{TLM conditioned on intro + pointer network extracted sentences} --- deep convolutional neural networks ( cnns ) are widely used in image recognition applications . however , they face a number of challenges . first , cnn models are computationally expensive to train and to perform . second , they are not easy to implement and can not exploit global features . to alleviate these issues , we present a method that leverages the correlation between semantic information present in feature maps and localization of an object of interest . the method leverages feature map activations for object localization . we demonstrate that this method is significantly faster than state - of - the - art in localization accuracy by a significant margin of up to 8 map on two standard data - sets with complex scenes , pascal voc 2012 and the much larger ms coco . \\
\hline
\textbf{Ground truth abstract} --- object localization is an important computer vision problem with a variety of applications . the lack of large scale object - level annotations and the relative abundance of image - level labels makes a compelling case for weak supervision in the object localization task . deep convolutional neural networks are a class of state-of-the-art methods for the related problem of object recognition . in this paper , we describe a novel object localization algorithm which uses classification networks trained on only image labels . this weakly supervised method leverages local spatial and semantic patterns captured in the convolutional layers of classification networks . we propose an efficient beam search based approach to detect and localize multiple objects in images . the proposed method significantly outperforms the state-of-the-art in standard object localization data - sets with a 8 point increase in map scores . \\
\hline
\end{tabular}
\end{center}
\label{tab:arxiv_random_set}
\end{table}
\subsection{T-SNE of learned word embeddings}
We visualize the word embeddings learned by our TLM model using t-sne. We find that words that are often associated with computer science are clustered in a different part of space when compared to words associated with physics. We use the arXiv REST API to find the submission category of each paper in the training set and then find the $\sim$300 most representative words for each category, using TF-IDF scores and plot them.

\end{document}